\documentclass{isprs}
\usepackage{setspace}
\usepackage{geometry}
\usepackage{epstopdf}
\usepackage[labelsep=period]{caption}
\usepackage[british]{babel} 
\usepackage[hang]{footmisc}
\usepackage{nicefrac}

\usepackage{amsmath,amssymb,amsfonts}
\usepackage{tikz}
\usepackage{pgfplots}
\pgfplotsset{compat=newest}
\usepackage{algorithm}
\usepackage{booktabs}
\usepackage{url}
\usepackage{xcolor}
\usepackage{attachfile}
\usepackage{algorithm}
\usepackage{algpseudocode}
\usepackage{stfloats}
\usepackage{fix-cm}

\begin{document}

\title{

Assessing the Impact of Fleet Size on Crowdsourced Mapping Using a Dissimilarity Measure}

\author{Marie-Ngoïe Badibanga Kalenda\textsuperscript{1}, Philippe Bonnifait\textsuperscript{2}, Marie-Anne Mittet\textsuperscript{3}}

\address{
	\textsuperscript{1}Université de Technologie de Compiègne, CNRS, Heudiasyc, Renault, Guyancourt, France - mbadiban@utc.fr\\
    \textsuperscript{2}Université de Technologie de Compiègne, CNRS, Heudiasyc, France - philippe.bonnifait@hds.utc.fr\\
	\textsuperscript{3}Renault, Guyancourt, France - marie-anne.n.mittet@ampere.cars \\
}

\abstract{
Accurate digital maps are essential for Advanced Driver Assistance Systems (ADAS) or Autonomous Driving (AD), providing critical information such as road geometry, traffic signs and speed limits required by safety functions including Intelligent Speed Assistance (ISA). Maintaining these map layers using traditional surveying methods is costly and difficult to scale. Crowdsourced approaches based on fleets provide a promising alternative for continuously validating and updating map information. However, the relationship between the number of contributing vehicles and the quality of the resulting map remains poorly understood.
To address this gap, this paper presents a simulation-based framework for evaluating crowdsourced traffic sign maintenance using a dissimilarity measure called $\text{GOSPA}_M$ (Generalized Optimal SubPattern Assignment for Maps), which combines localization errors with detection performance by accounting for False Positives (FP) and False Negatives (FN). The proposed system models multi-vehicle observations with representative sensor noise, detection errors, and semantic recognition uncertainties. Observations from multiple vehicles are aggregated using spatial clustering and semantic filtering to estimate traffic sign locations.
Using simulated trajectories generated from data carried out by an experimental vehicle in an area containing ground-truth traffic signs, we assess the influence of fleet size on the performance of crowdsourced mapping. The number of vehicles ranges from 5 to 50, and performance is analyzed using standard evaluation metrics which are compared to the $\text{GOSPA}_M$. 
The results show that $\text{GOSPA}_M$ can be used to effectively assess the quality of crowdsourced mapping, such as the contributions made by the first vehicles or the improvements made by numerous vehicles.
}


\maketitle
\section{Introduction}\label{Introduction}

High Definition (HD) maps containing accurate traffic sign information are critical for autonomous driving systems, which rely on these digital representations for safe navigation and regulatory compliance \cite{wijaya2024high}. Traditional map maintenance using high-precision Mobile Mapping Systems (MMS) is increasingly completed by crowdsourced strategies leveraging connected vehicle fleets \cite{kim_updating_2021,kim_hd_2021,kim_crowd-sourced_2018}. While crowdsourcing offers scalability and cost-effectiveness, it introduces significant challenges related to sensor variability, localization uncertainty, and partial feature visibility \cite{wong_mapping_2020}.

Traffic signs present unique challenges for crowdsourced validation. Unlike continuous features such as lane markings, traffic signs are discrete point features that require both accurate localization and correct semantic classification (Table \ref{tab:traffic_signs_comp_map}). Misclassifying a stop sign as a yield sign can have severe safety implications that cannot be corrected through simple spatial aggregation. Furthermore, traffic signs exhibit high variability in detection characteristics. Although they are generally easier to detect because of their distinctive shapes, they can generate more false positives from similar objects such as poles or billboards.

In real-world situations, changes in roads frequently occur due to road works, infrastructure modifications, or traffic signs updates intended, for example, to adapt speed limits. These changes are generally restricted to a specific road area. In such localized areas, the map must be updated, and crowdsourcing represents a promising solution for rapidly updating map information. This constitutes the problem addressed in this paper.

However, one fundamental question remains insufficiently explored: how many vehicles are required for a crowdsourced mapping system to produce reliable map updates ? While increasing the number of contributing vehicles intuitively improves map quality, the relationship between fleet size and mapping performance remains poorly understood. Evaluating this relation requires appropriate metrics and measures capable of jointly capturing localization accuracy and detection completeness. In this work, we explore the use of a global dissimilarity measure named  $\text{GOSPA}_M$ \cite{badibanga_kalenda_vector_2025}, which combines localization errors with penalties for false positives and false negatives into a single measure of map quality.

This paper investigates the influence of vehicle fleet size on the quality of crowdsourced traffic sign mapping. To enable controlled experimentation, we developed a fleet simulation framework that models multi-vehicle observations of traffic signs under realistic sensing uncertainties, including localization noise, false positives, false negatives, and semantic recognition errors.

The main contributions of this work are : 
\begin{enumerate}
   \item \textbf{Evaluation of aggregation strategies}: A comparison of several spatial clustering algorithms combined with a semantic classification of the traffic signs. 

    \item \textbf{Development of a simulation framework for controlled evaluation}: A simulator capable of generating realistic traffic sign detections from vehicle-mounted cameras while modeling configurable sensor errors.
   
   \item \textbf{Analysis of fleet size impact}: A systematic evaluation of how the number of contributing vehicles influences the quality of crowdsourced traffic sign mapping using both classical metrics and the $\text{GOSPA}_M$. 
    
\end{enumerate}

The remainder of this paper is organized as follows: Section \ref{Related work} reviews the related work, Section \ref{Crowdsourcing Architecture} explains the crowdsourcing architecture, Section \ref{Simulation Methodology} details the methodology: detection simulation and aggregation methodology. Section \ref{Evaluation metrics} details the metrics, Section \ref{Results} reports experimental results including the clustering comparison and the influence of the fleet size and Section \ref{Conclusion} concludes.

\begin{table}
    \centering
    \begin{tabular}{c c}
        \includegraphics[width=0.30\linewidth]{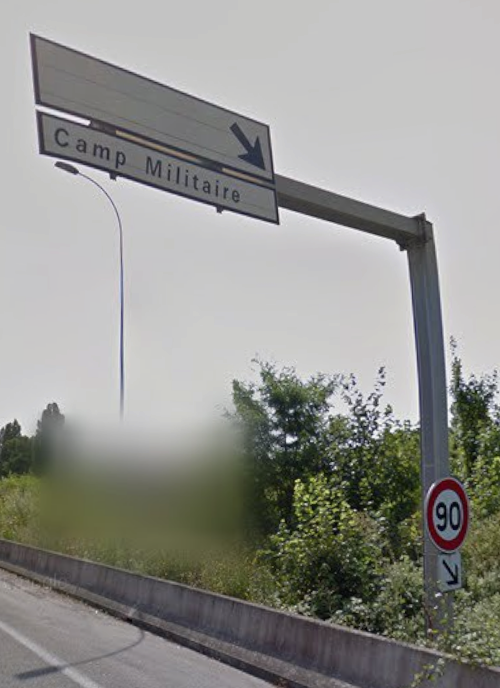} &
        \includegraphics[width=0.25\linewidth]{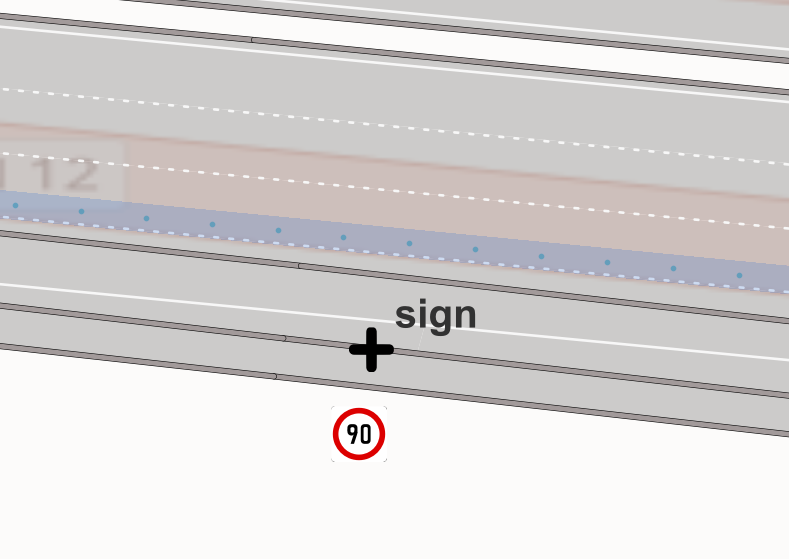} \\

        \includegraphics[width=0.45\linewidth]{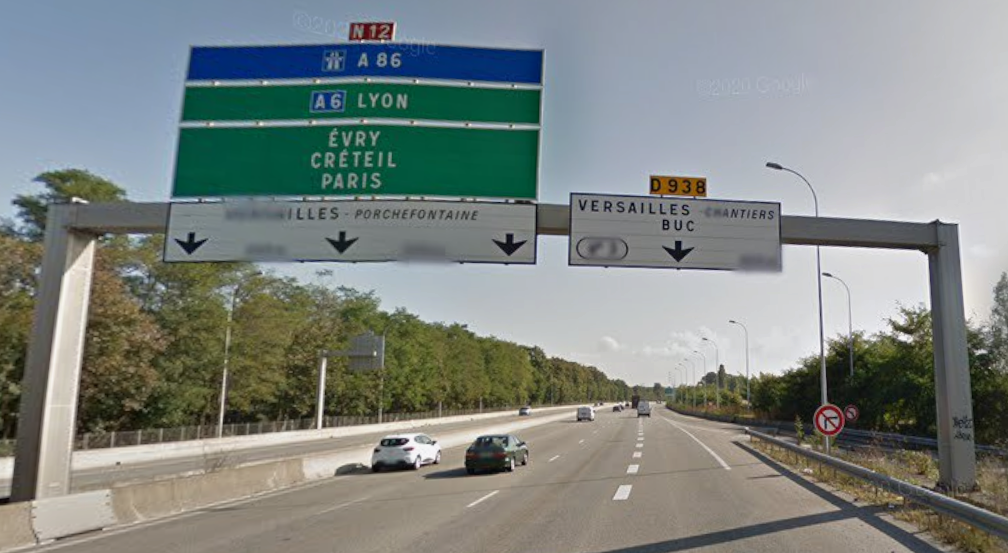} &
        \includegraphics[width=0.25\linewidth]{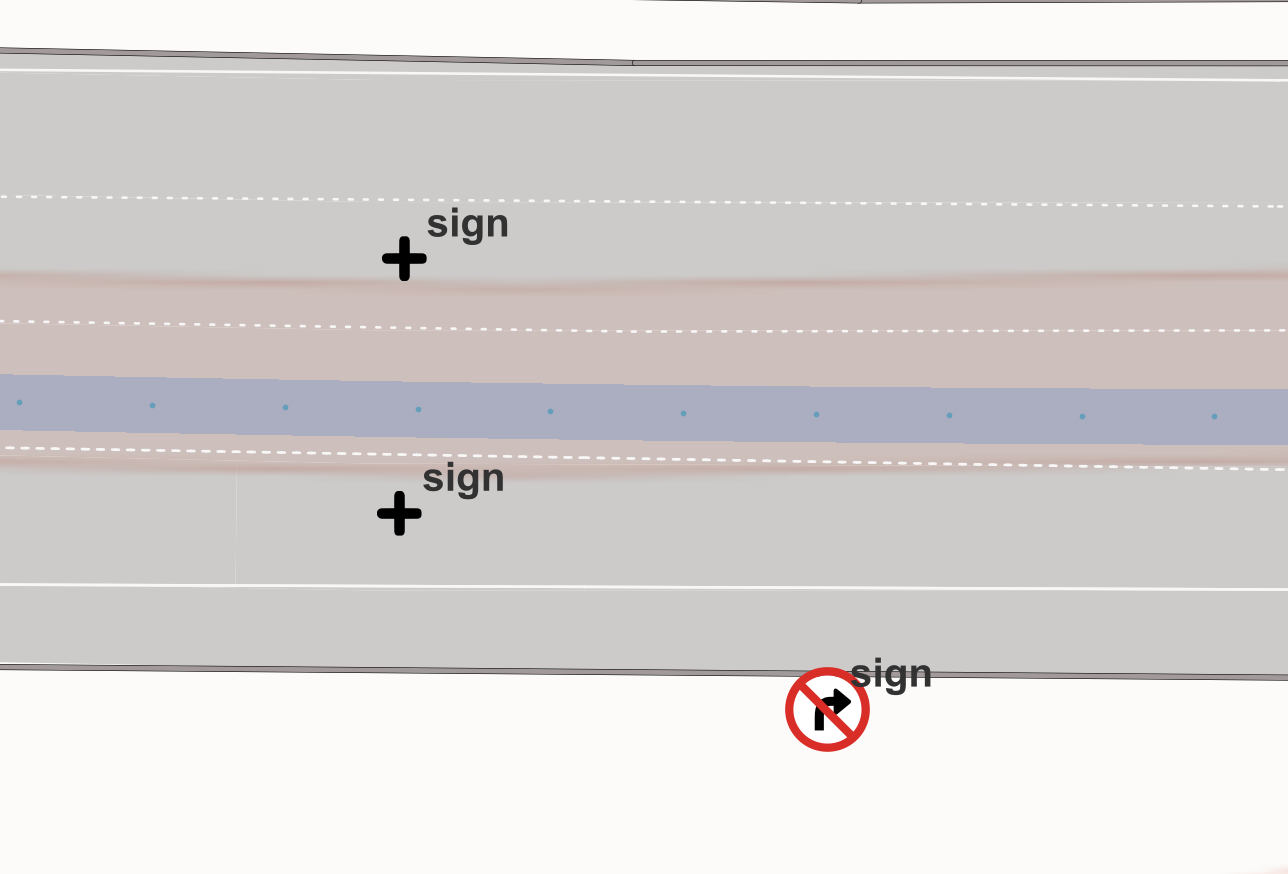} \\
    \end{tabular}
    \caption{Examples of traffic signs in an HD Map. The ground projection of the center of the sign is mapped.}
    \label{tab:traffic_signs_comp_map}
\end{table}

\section{Related work}\label{Related work}

HD map updating through crowdsourced vehicle data has become a major research direction as large-scale manual surveying is costly and difficult to maintain over time. Recent surveys highlight crowdsourcing as a key solution for ensuring both map freshness and accuracy, especially for autonomous driving applications that depend on precise lane-, road-, and landmark-level information \cite{guo_review_2024}. Crowdsourced update pipelines typically include data collection from heterogeneous sensors, feature extraction, change detection, and map fusion, with each stage presenting significant challenges due to noise, sparse sampling, and variations in sensor quality.

Recent work has explored various aspects of crowdsourced mapping. Feature-level map matching using graph-based SLAM has been proposed \cite{pannen_how_2020}, and advanced filtering algorithms such as DBSCAN have been used to remove noisy observations \cite{moawad_offline_2025}. End-to-end crowdsourced 3D mapping systems have been developed for autonomous vehicles \cite{dabeer_end--end_2017}, and semantic edge mapping from crowdsourced data has shown promise for localization \cite{herb_crowd-sourced_2019}. HD map generation from noisy multi-route fleet data has also been addressed using the Expectation-Maximization algorithm \cite{immel_hd_2023}, while large-scale updates of lane-level HD maps have been demonstrated using crowdsourced bus and taxi data \cite{cho_frequent_2023}. 

Several methods have explored traffic sign updating using vehicle-based sensing. Incremental map update strategies fuse observations collected over multiple journeys to detect new or modified signs and refine their positions, demonstrating improved localization accuracy through optimized fusion techniques \cite{hu_incremental_2025}. Other full-stack systems propose end-to-end mapping architectures using low-cost sensors such as consumer-grade cameras and GPS receivers, achieving impressive sub-meter accuracy for traffic sign reconstruction through multi-route triangulation, clustering, and bundle adjustment \cite{dabeer_end--end_2017}. These approaches validate the feasibility of large-scale HD map construction from crowdsourced data but often rely on deterministic pipelines with fixed sensor characteristics.

Beyond traffic signs, crowdsourced data have been used to extract broader road network structures. Intersection-first reconstruction methods based on large taxi trajectory datasets can recover road geometry and topology even under low-frequency sampling conditions, enabling scalable road network generation \cite{zhang_intersection-first_2019}. Similarly, frequent lane-level map updates have been demonstrated using dense vehicle fleets such as buses and taxis, capturing environmental changes through pose correction, observation clustering, and landmark classification \cite{cho_frequent_2023}. However, these methods primarily focus on continuous or linear features (e.g., lanes, road edges) rather than discrete punctual landmarks such as traffic signs.

Recent work has underscored the importance of robust multi-vehicle fusion and advanced uncertainty handling for crowdsourced mapping. A comprehensive review shows that, although numerous pipelines exist for HD map updating, challenges remain in modeling heterogeneous sensors, mitigating detection errors, and designing aggregation strategies that scale effectively with fleet size \cite{guo_review_2024}. These limitations are particularly relevant for traffic signs, for which false positives and false negatives have strong impacts on map reliability, and where semantic consistency is essential.

Overall, existing approaches demonstrate the potential of crowdsourced mapping but leave open fundamental questions regarding how map quality scales with the number of contributing vehicles, how multi-vehicle traffic sign detections should be optimally combined, and how such systems should be evaluated using metrics and measures that jointly account for localization and detection performance. These open questions motivate the simulation-based framework proposed in this study, which explicitly models sensor uncertainty, detection errors, and multi-vehicle aggregation to quantify the relationship between fleet size and traffic sign mapping performance.

\section{Crowdsourcing Architecture}\label{Crowdsourcing Architecture}

\begin{figure*}[!t]
    \centering
    \includegraphics[width=0.8\textwidth]{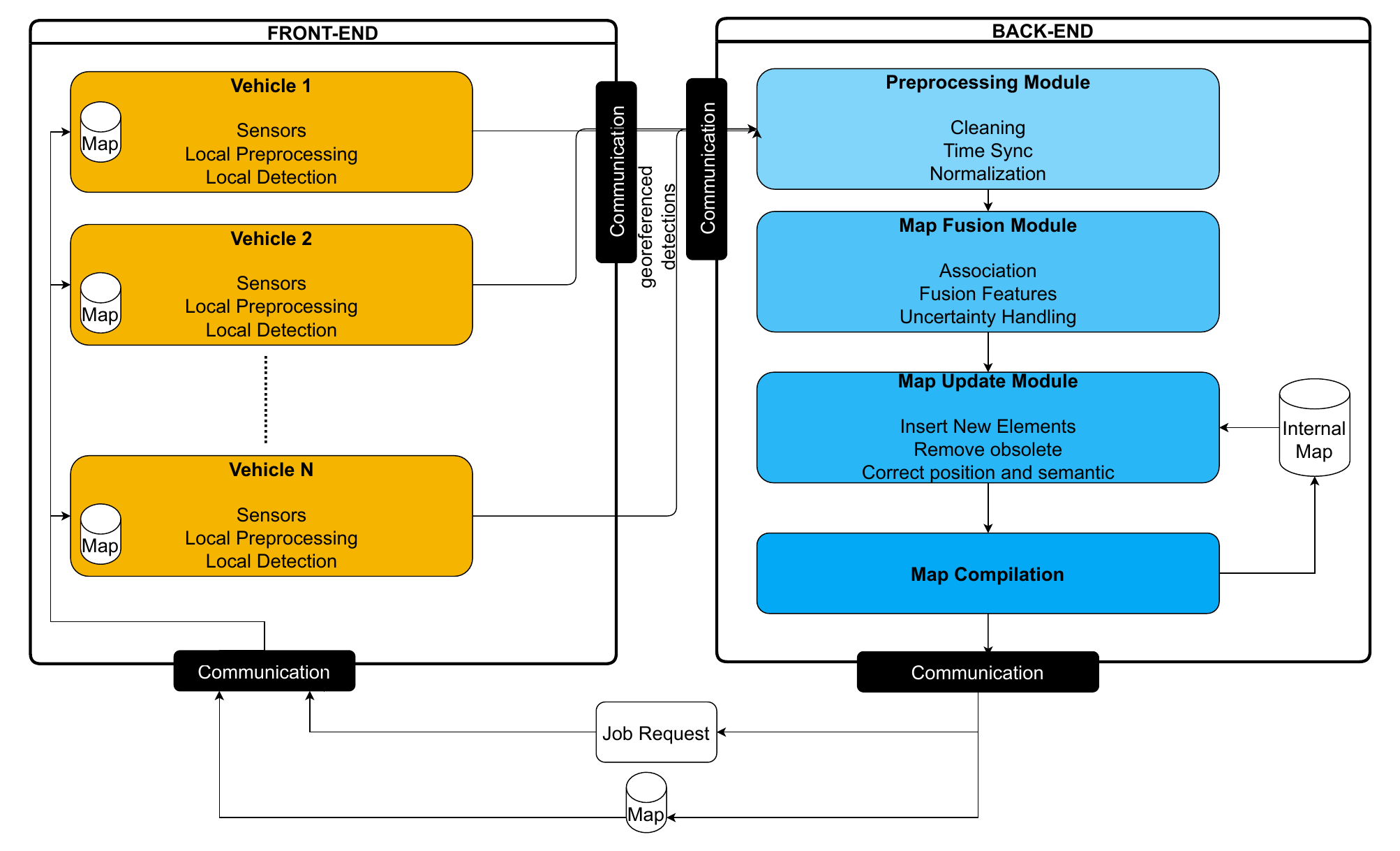}
   \caption{System diagram: Front-End (Vehicles) feeding the Back-End (Cloud) through data harvesting. 
   When a change is reported in a geographical area, a “Job Request” is sent to vehicles passing through that area, asking them to upload their data to the back-end system in charge of updating the map. 
   }
\label{fig:architecture}
\end{figure*}

The proposed crowdsourced mapping framework follows a two-stage architecture composed of a vehicle-side front end and a cloud-based back-end, as illustrated in Figure \ref{fig:architecture}. The objective of this architecture is to collect traffic sign observations from multiple vehicles and aggregate them in order to maintain and update the corresponding tile of the map used by the vehicles.

The proposed fleet simulation implements the crowdsourcing architecture illustrated in Figure \ref{fig:architecture}.

\textbf{Front-End}: 
On the front-end, multiple vehicles equipped with perception sensors detect traffic signs while traveling along the road network. Each vehicle performs local preprocessing such as temporal tracking for position refinement. The vehicle generates traffic sign observations associated with timestamps, spatial coordinates, and semantic attributes. These georeferenced observations are map-matched to the onboard map. If a change is detected or a job request (i.e., a request of new detections) is ongoing in the tile (i.e., a portion of the full map), the data are transmitted through a communication layer to the back-end infrastructure.

\textbf{Back-End}: 
The back-end first applies preprocessing operations, including data cleaning, timestamping, and coordinate system normalization. The observations are then processed by a map fusion module that associates detections coming from different vehicles and handles uncertainty in multi-vehicle observations.
Finally, a map update module integrates the aggregated detections into the internal map representation by inserting new landmarks, removing outdated elements, and updating the location and/or semantics of elements that have already been mapped. Metric and quality measures calculations  are then performed on the updated map, including the computation of $\text{GOSPA}_M$ and conventional metrics (precision, recall and F1-score). Based on these, the updated map can then be compiled. As long as the confidence in the new values remains insufficient and the map quality has not reached the required level, the job request remains active and is sent to vehicles in the relevant map tile. Consequently, every vehicle entering this tile transmits its data to the cloud-based back-end. Conversely, once the required quality threshold is reached, the updated map is distributed through the communication layer to all vehicles. Their onboard maps are then updated, and the job request is terminated.

\section{Simulation Methodology}\label{Simulation Methodology}

\begin{figure}
      \centering
    \includegraphics[width=1\linewidth]{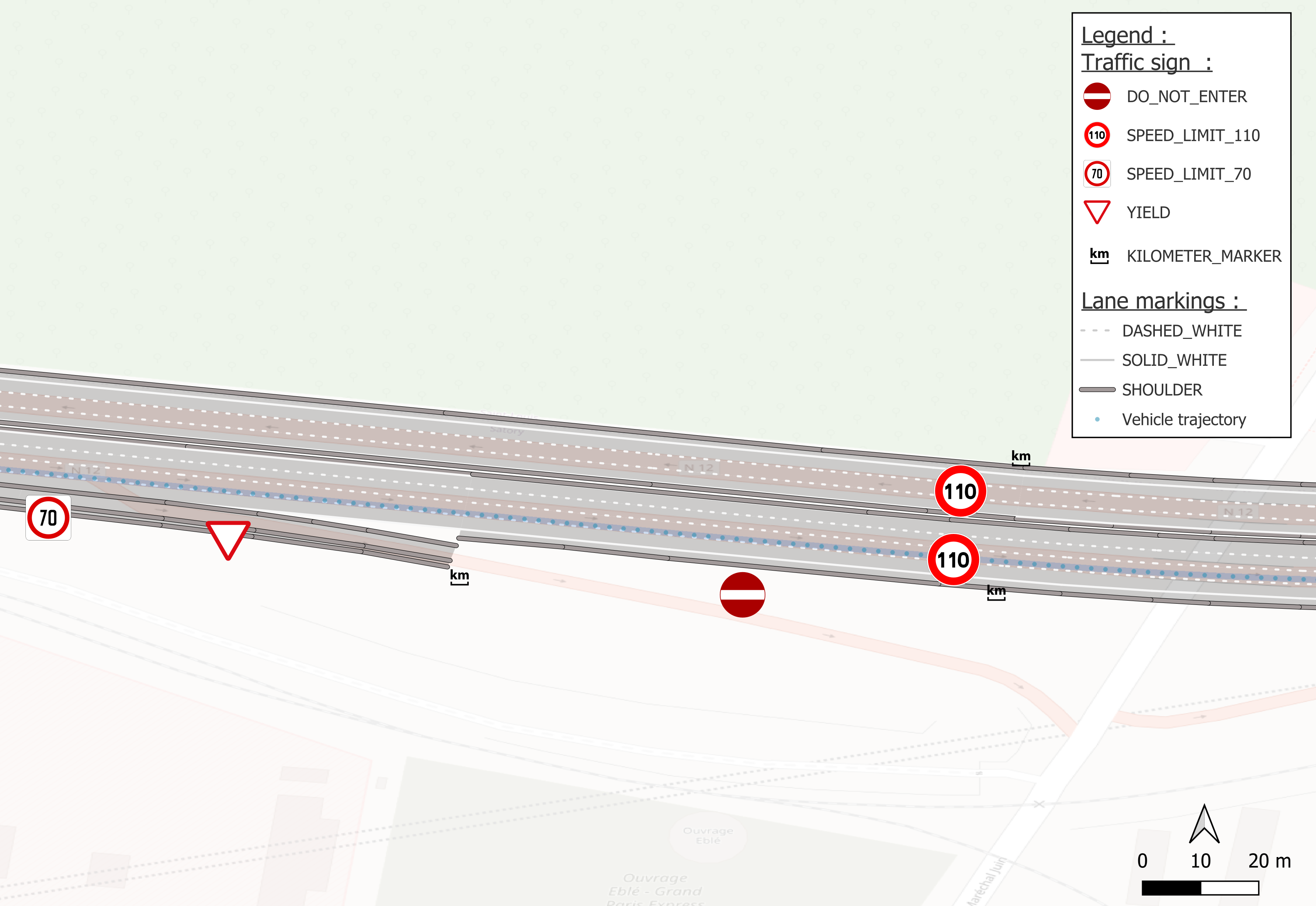}
    \caption{Zoom-in view of the reference map used for the tests}
    \label{fig:groundtruth}
\end{figure}

To reproduce realistic conditions within a controlled experimental environment, traffic sign detections are simulated based on a reference HD map containing ground-truth landmark positions and semantic classes (see Fig.~\ref{fig:groundtruth}). 
We extracted a section of an actual vehicle path driven by a Renault experimental vehicle, which was tracked with high accuracy using a RTK GNSS receiver. 
Using this sequence as a basis, we generated through simulation a fleet of vehicles following approximately the same route. 
As vehicles follow the predefined trajectory with small variations representative of real-world inter-vehicle variability, traffic signs located within the sensor detection range and field of view are considered observable and may generate detection events. In the proposed crowdsourced mapping pipeline, traffic sign observations are generated on the vehicle side of the system (front-end), as illustrated in Fig. \ref{fig:architecture}. We consider the scenario of an active "Job Request", meaning that the map must be fully updated in the corresponding area. Each simulated vehicle generates georeferenced detections of traffic signs affected by realistic sensing uncertainties, including localization noise (e.g., GNSS errors), false positive detections (ghost detections), false negatives (missed detections), and semantic recognition errors. Our simulation framework introduces configurable detection errors in order to approximate the behavior of onboard perception algorithms. These detections are tracked across successive camera frames in order to reproduce the tracking process performed by smart cameras. When a sign leaves the sensor's field of view, the tracking result is sent to the back-end. These different error sources are described in the following sections.

Since multiple vehicles may detect the same traffic sign, the resulting observations must be aggregated by the back-end in order to estimate the final location of the signs. The aggregation process therefore relies on two key elements: the semantic class and the location of the detected sign. In order to determine a suitable aggregation strategy for the crowdsourced mapping pipeline, a comparison of clustering methods is conducted later in this paper.

To investigate the influence of fleet size on mapping quality, the number of contributing vehicles can be selected. For each configuration, multiple simulation runs are performed, and traffic sign mapping quality is evaluated using performance metrics and quality measures.

\subsection{Detection Simulation (front-end)}
Objects are detected if they fall within the sensor's field of view:
\begin{equation}
\text{detected} = \begin{cases}
1 & \text{if } d < r \text{ and } |\theta| < \frac{\text{FOV}}{2} \\
0 & \text{otherwise}
\end{cases}
\end{equation}
where $d$ is the distance to the object, $r$ is the sensor range, and $\theta$ is the azimuth (bearing) angle from the sensor heading.

In practice, front cameras are capable of detecting and recognizing landmarks at distances up to 100 meters. 
A typical semantic recognition rate is of 95 \%, a false positive rate of around 2 \%, and a false negative rate of around 10 \%. 
Therefore, these values were selected for the simulation.

\subsubsection{Traffic Sign Detection Generation}

Real sensors exhibit two distinct types of measurement errors that are  modeled separately to achieve realistic simulation of vehicle pose estimation uncertainties.

\textbf{Sensor-setting error} ($\epsilon_{\text{sensor}}$): Systematic errors affecting all measurements from a sensor, including mounting position errors and calibration drift. This noise is sampled once per sensor and applied to all detections during a simulation run:
\begin{equation}
\epsilon_{\text{sensor}} \sim \mathcal{N}(0, \sigma_{\text{sensor}}^2)
\end{equation}

\textbf{Measurement noise} ($\epsilon_{\text{object}}$): Random errors varying between measurements due to factors such as environmental conditions. This noise is sampled independently for each detected traffic sign:
\begin{equation}
\epsilon_{\text{object}} \sim \mathcal{N}(0, \sigma_{\text{object}}^2)
\end{equation}

The measured distance to a traffic sign combines both noise sources:
\begin{equation}
d_{\text{measured}} = d_{\text{true}} + \epsilon_{\text{sensor}} + \epsilon_{\text{object}}
\end{equation}

\textbf{Angle noise ($\theta$)}: An analogous noise model is used for angular measurements, with both sensor-setting error ($\theta_{\text{sensor}}$) and measurement noise ($\theta_{\text{measured}}$):
\begin{equation}
\theta = \theta_{\text{true}} + \epsilon_{\theta}
\end{equation}
where $\epsilon_{\theta} \sim \mathcal{N}(0, \sigma_{\theta}^2)$ .

The 2D position error is calculated similarly for sensor-setting and measurement noise through polar-to-Cartesian conversion in the vehicle frame, as illustrated below for the measured position:
\begin{equation}
\begin{aligned}
x_{\text{measured}} &= d_{\text{measured}} \cos(\theta_{\text{measured}}) \\
y_{\text{measured}} &= d_{\text{measured}} \sin(\theta_{\text{measured}})
\end{aligned}
\end{equation}

This two-level approach produces representative positioning errors of a few meters, consistent with automotive-grade sensors.

\textbf{Semantic Misclassification}: Semantic recognition is applied in the front-end with configurable recognition rates (e.g., 90\% for traffic signs). When recognition fails, a random valid alternative type is chosen based on Alg. \ref{Semantic_Recognition}, simulating misclassifications caused by:
\begin{itemize}
\item Similar visual appearance (e.g., circular speed limit signs)
\item Partial occlusion obscuring key features
\item Poor viewing angles reducing sign visibility
\item Weather effects (rain, fog, or snow) degrading image quality
\end{itemize}

This modeling framework enables the evaluation of the impact of semantic recognition errors on traffic sign validation quality, which is critical for safety-critical applications where misclassifying a STOP sign as a YIELD sign has severe implications \cite{wijaya_crowdsourced_2022}.

\begin{algorithm}

    \begin{algorithmic}[1]
    \For{each detected object with a semantic type}
        \State $r \gets$ random() \Comment{Uniform distribution over [0,1]}
        \If{$r < r_{\text{semantic}}$}
            \State Keep the correct semantic type
            \State Mark as semantic\_correct = true
        \Else
            \State Choose a random valid alternative type
            \State Mark as semantic\_correct = false
        \EndIf
        \State Store reference\_semantic, detected\_semantic
    \EndFor
    \end{algorithmic}
    \caption{Semantic Recognition}
    \label{Semantic_Recognition}
\end{algorithm}

\subsubsection{Traffic Sign False Positives and False Negatives }
~

Traffic signs  exhibit distinct detection characteristics compared to continuous road features such as lane markings. They are easier to detect due to their distinctive shapes and retroreflective materials but generate more false positives from similar objects such as poles, billboards, or building features. 

False negatives are applied after cone-based detection by randomly removing detected signs with probability equal to the FN rate, simulating occlusion by vehicles or vegetation, poor lighting conditions, viewing angle limitations, and sensor blind spots. False positives are generated using a Poisson process:
\begin{equation}
N_{\text{FP}} \sim \text{Poisson}(\lambda), \quad \lambda = \text{FP\_rate} \times A_{\text{detection}}
\end{equation}
where $A_{\text{detection}} = \frac{1}{2} r^2 \theta_{\text{FOV}}$ is the detection cone area. False positive positions are uniformly distributed within the cone, simulating the misdetection of poles, building features, and other vertical structures.

\subsubsection{Temporal Tracking and Fusion}
The system implements temporal tracking at the front-end to refine traffic sign positions across multiple frames within a vehicle run. 

Across successive frames, spatially-fused positions are associated and averaged to reduce measurement noise.

Frame-to-frame association uses the Hungarian algorithm with a cost matrix based on Euclidean distance in the ENU (East-North-Up) frame. Traffic signs within a 2.0 m gating distance are considered potential matches. The estimated position P is computed as:
\begin{equation}
\mathbf{p} = \frac{1}{F} \sum_{f=1}^{F} \mathbf{p}_{frame,f}
\end{equation}

where $F$ is the number of frames in which the sign was detected while remaining within the sensor's field of view.

Position uncertainty is quantified using the standard deviation:
\begin{equation}
\sigma_p = \sqrt{\frac{1}{F-1} \sum_{f=1}^{F} (p_f - \bar{p})^2}
\end{equation}
where $p_f$ denotes the coordinates of the detected object in frame $f$ and $\bar{p}$ is the mean position.

\subsection{Multi-Vehicle aggregation (back-end)}
In a crowdsourced mapping system, multiple vehicles may observe the same traffic sign at different times and from different viewpoints. These independent observations must therefore be aggregated in order to estimate the position and semantic class of the corresponding map landmark. 
The aggregation process consists of two main steps. 

\subsubsection{Semantic filtering} A key stage in the processing is the semantic management strategy applied to multi-vehicle detections, which groups detections by semantic type (e.g., SPEED LIMIT, YIELD). This strategy can be applied before or after spatial clustering. 


\subsubsection{Spatial Clustering}
Traffic sign detections from multiple vehicles are clustered using spatial aggregation \cite{chetouane_application_2022}. Clustering is a fundamental unsupervised learning task, and numerous surveys highlight the diversity of clustering approaches and their limitations \cite{wani_comprehensive_2024,xu_comprehensive_2015,farahnakian_comprehensive_2023}. Four clustering algorithms were evaluated: DBSCAN (density-based), K-Means (centroid-based), K-Medoids (medoid-based, robust to outliers), and HDBSCAN (hierarchical density-based, automatic cluster selection). Euclidean distance is computed in  ENU Cartesian coordinates after converting geodetic coordinates (e.g., WGS84).

\paragraph{DBSCAN (Density-Based Spatial Clustering of Applications with Noise)}
DBSCAN groups points that are closely packed together based on a neighborhood radius $\varepsilon$ (e.g. $\varepsilon$ = 2*gating distance = 4m). It does not require specifying the number of clusters and naturally handles noisy points, making it widely used in spatial applications and anomaly detection \cite{wani_comprehensive_2024,farahnakian_comprehensive_2023}. 
The cluster structure depends entirely on the density defined by the $\varepsilon$ neighborhood. Cluster centroids are computed as weighted averages in the local ENU frame and then converted back to geodetic coordinates.

\paragraph{K-Means}

K-means partitions detections into $K$ clusters by iteratively minimizing within-cluster variance. 
K-Means uses the algorithm's computed mathematical means as cluster centers (not weighted centroids), which makes it sensitive to outliers (a well-known limitation emphasized in general clustering surveys \cite{xu_comprehensive_2015,wani_comprehensive_2024}). A single false positive detection far from the true cluster can shift the centroid significantly.

\paragraph{K-Medoids (PAM --- Partitioning Around Medoids)}

K-Medoids is similar to K-Means but constrains cluster centers to be actual data points (medoids), providing greater robustness to outliers. 
Because cluster centers are actual detections, K-Medoids avoids the centroid-shift problem of K-Means. However, it is computationally more expensive ($O(N^2)$ per iteration) \cite{xu_comprehensive_2015}.

\paragraph{HDBSCAN (Hierarchical Density-Based Spatial Clustering)}

HDBSCAN extends DBSCAN by building a hierarchy of clusters at varying density levels and selecting the most persistent clusters, thereby removing the need to explicitly set $\varepsilon$ or $K$. Hierarchical density-based clustering is particularly effective in mixed-density environments \cite{wani_comprehensive_2024}.  
HDBSCAN tends to produce more clusters than DBSCAN because it identifies density-based structure at multiple scales. In our experiments, this led to over-segmentation when combined with semantic-first filtering (many small clusters per semantic type), but performed competitively without semantic-first filtering.

\section{Evaluation metrics and quality measures }\label{Evaluation metrics}
To evaluate the quality of a crowdsourced map, we use a reference map, namely the map used by the front end of our simulator. 

\subsection{Usual metrics}
For a given tile of the map to be evaluated, the matched points are labeled TP and the others FP. The unmatched points of the reference map are FN.
Comparing the evaluated map to the reference map, it is then possible to calculate the "Precision":
\begin{equation}\label{Precision}
\text{Precision}=\frac{TP}{TP + FP}
\end{equation}   
It represents the proportion of the map features that exist in the real world.

Equivalently, "Recall" is defined as:
  \begin{equation}\label{Recall}
    \text{Recall} =\frac{TP}{TP + FN}
    \end{equation}
It represents the proportion of the map features that exist in the real world.
A map that has a high recall is a map that is complete (or almost complete). 
   
The F1-score is also often used to combine Precision and Recall  in a single number. It corresponds to the harmonic mean of Precision and Recall:
\begin{equation}
F_{1} = 2 \cdot \frac{\text{Precision} \cdot \text{Recall}}{\text{Precision} + \text{Recall}} \in [0,1]
\label{F1}
\end{equation}


A map that has a high F1-score has a good compromise between precision and recall.

Consider now only the TPs, i.e. the set of the features ${x_k, y_p}$ map-matched with the reference. 
The Mean Square Error  of the matched features ($\text{MSE}_M$), is defined as:
\begin{equation}\label{RMSE_M}
\text{MSE}_M=\frac{\sum_{k,p \in \gamma} (x_k, y_p)^2}{TP},   
\end{equation}
where $\gamma$ denotes the set of matched pairs between the map and the reference.

This metric quantifies the average metric error (also called metric accuracy) of the features of the map to be evaluated but it doesn't handle the FN and FP values (i.e. Precision or Recall). 

These metrics are classically used to evaluate vector maps, but they do not provide a single numerical value for the overall quality of a map. 

\subsection{\texorpdfstring{$\mathrm{GOSPA}_M$}{GOSPA M}}
$\text{GOSPA}_M$ is an adaptation of the GOSPA \cite{rahmathullah_generalized_2017} commonly used to evaluate the performance of perception and tracking systems.
\begin{equation}\label{GOSPA4}
\text{GOSPA} = \left( \sum_{k,p \in \gamma} d(x_k, y_p)^{2} + \frac{c^2}{2} (FN + FP) \right)^{\frac{1}{2}} 
\end{equation}

$\text{GOSPA}_M$ combines localization accuracy with detection completeness into a single measure. 
If $TP\neq 0$, $\text{GOSPA}_M$ is defined in meters as follows \cite{badibanga_kalenda_vector_2025}:
\begin{equation}\label{GOSPA_M}
\text{GOSPA}_M  = \frac{\text{GOSPA}}{\sqrt{ TP }} 
\end{equation}
By normalizing by $TP$ (rather than by $N$, the total number of elements in the map), one obtains an expression in which $\text{MSE}_M$ appears explicitly
\begin{equation}
\text{GOSPA}_M = \sqrt{\text{MSE}_M + \tfrac{c^2}{2} \left( \tfrac{FN + FP}{TP} \right)}, \, TP \neq 0,
\end{equation}
where $c$\ is the gating distance of the map-matching (called "cut‑off" distance in the classical GOSPA). 

$\text{GOSPA}_M$ therefore combines the position accuracy of the matched elements ($\text{MSE}_M$) with a penalty applied to account for phantom or missing elements.  More precisely, the term  $\nicefrac{(FN + FP)}{TP}$ refers to the ratio of incorrect or missing elements contained in the map to those that are correct. Its value can become very high for very poor-quality maps. 

$\text{GOSPA}_M$ provides a way of assessing the quality of a map using a single indicator.
It can be seen as a measure of dissimilarity between two maps. The lower its value, the more similar the maps are (with 0 corresponding to two perfectly identical maps). When $\text{GOSPA}_M$  is computed between a given map $M$ and a reference, it quantifies the quality of the map $M$.

\section{Results}\label{Results}

The experiments were conducted using a road segment extracted from a HD map provided by a commercial map supplier based on a real vehicle trajectory acquired by the team. The urban section considered in this section is approximately 1.2\,km long and contains different layers such as a traffic-sign layer and a lane-geometry layer. The reference dataset contains 33 ground-truth traffic signs.
This case study, which focuses on a local area, is a good example of the kind of problem we are considering in this paper. 

\subsection{Clustering strategy selection}
\hyphenation{approaches}

To determine an appropriate aggregation strategy for crowdsourced traffic sign detections, the presented clustering approaches were evaluated by applying the semantic clustering strategy before or after. 

Table~\ref{tab:clustering_comparison} summarizes the results obtained for the different clustering algorithms evaluated on the dataset containing 33 ground-truth traffic signs, for a number of 50 vehicles. 
Since K-Means and K-Medoids require the number of clusters, we provided these algorithms with this value in order to evaluate them under the most favorable conditions. 

Among the tested approaches, DBSCAN combined with semantic-first grouping achieves the best overall performance, reaching an F1-score of 90.6\% with high precision (93.5\%) and recall (87.9\%), as well as the lowest $\text{GOSPA}_M$ value (0.756\,m).
K-Means provides competitive performance (F1=86.2\%, \\ $\text{GOSPA}_M$=0.928\,m) but the performance gap is still significant. HDBSCAN achieves high precision (92.0\%) but suffers from lower recall (69.7\%). K-Medoids performs significantly worse due to the irregular spatial distribution of traffic sign detections.

This comparison shows that both the F1-score and $\text{GOSPA}_M$ provide meaningful criteria for comparing aggregation methods. 


The comparison also highlights the importance of semantic-first grouping. Without this step, spatial clustering tends to merge nearby detections corresponding to different traffic sign types, a situation that frequently occurs in urban environments where multiple signs may be mounted on the same structure.


\begin{table*}[htbp]
\centering

\begin{tabular}{cccccc}
\toprule
Algorithm & Sem.-First & Prec. (\%) & Rec. (\%) & F1 (\%) & $\text{GOSPA}_M$ (m) \\
\midrule
\textbf{DBSCAN}   & \textbf{Yes} & \textbf{93.5} & \textbf{87.9} & \textbf{90.6} & \textbf{0.756331172} \\
K-Means           & Yes          & 87.5          & 84.8          & 86.2          & 0.927746939 \\
K-Means           & No           & 86.7          & 78.8          & 82.5          & 1.073600069 \\
HDBSCAN           & Yes          & 92.0          & 69.7          & 79.3          & 1.080425661 \\
HDBSCAN           & No           & 85.2          & 69.7          & 76.7          & 1.201992465 \\
DBSCAN            & No           & 80.8          & 63.6          & 71.2          & 1.314174608 \\
K-Medoids         & Yes          & 73.7          & 42.4          & 53.8          & 1.981584718 \\
K-Medoids         & No           & 42.9          & 18.2          & 25.5          & 3.588900265 \\
\bottomrule
\end{tabular}
\caption{Complete Clustering Algorithm Comparison (33 GT, sorted by $\text{GOSPA}_M$ score).}
\label{tab:clustering_comparison}
\end{table*}

\subsection{Influence of fleet size on mapping quality}
Using the clustering configuration selected in the previous section (semantic-first grouping followed by DBSCAN), we evaluated the influence of fleet size on the quality of the reconstructed traffic sign map. The number of contributing vehicles was progressively varied from 5 to 50 vehicles. For each configuration, the experiment was repeated ten times in order to account for stochastic variations in the simulated detection process. The reported results correspond to the mean values obtained across the ten runs, with error bars on the $\text{GOSPA}_M$ representing the standard deviation.

\begin{figure}[!t]
    \centering
    \begin{tikzpicture}
        \begin{axis}[
            xlabel={Number of vehicles},
            ylabel={F1 Score},
            ymin=0.70, ymax=0.95,
            grid=both,
            width=7.5cm,
            height=5.5cm
        ]
        \addplot+[mark=*, color=green]
        coordinates {
            (5,0.7495)
            (10,0.8345)
            (15,0.8677)
            (20,0.8935)
            (25,0.9149)
            (30,0.9149)
            (35,0.9255)
            (40,0.9271)
            (45,0.9143)
            (50,0.9300)
        };
        \end{axis}
    \end{tikzpicture}
    \caption{F1 score as a function of vehicle count.}
    \label{fig:f1_vs_vehicles}
\end{figure}
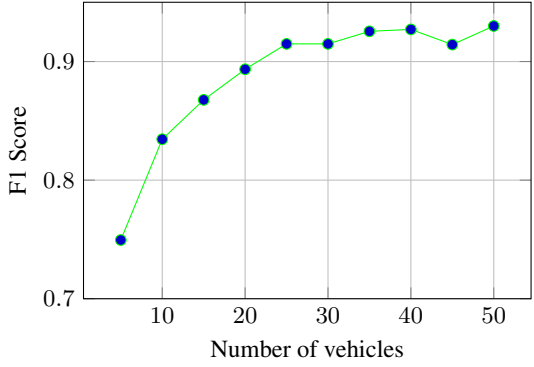

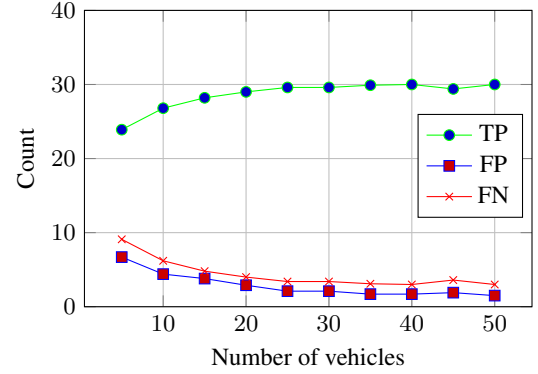
\begin{figure}[!t]
    \centering
    \begin{tikzpicture}
        \begin{axis}[
            xlabel={Number of vehicles},
            ylabel={Count},
            ymin=0, ymax=40,
            legend style={at={(0.98,0.3)}, anchor=south east},
            grid=both,
            width=7.5cm,
            height=5.5cm
        ]
        
        \addplot+[mark=*, color=green]
        coordinates {
            (5,23.9) (10,26.8) (15,28.2) (20,29.0)
            (25,29.6) (30,29.6) (35,29.9)
            (40,30.0) (45,29.4) (50,30.0)
        };
        \addlegendentry{TP}

        \addplot+[mark=square*, color=blue]
        coordinates {
            (5,6.7) (10,4.4) (15,3.8) (20,2.9)
            (25,2.1) (30,2.1) (35,1.7)
            (40,1.7) (45,1.9) (50,1.5)
        };
        \addlegendentry{FP}

        \addplot+[mark=x, color=red]
        coordinates {
            (5,9.1) (10,6.2) (15,4.8) (20,4.0)
            (25,3.4) (30,3.4) (35,3.1)
            (40,3.0) (45,3.6) (50,3.0)
        };
        \addlegendentry{FN}
        
        \end{axis}
    \end{tikzpicture}
    \caption{Detection statistics (TP/FP/FN) as a function of vehicle count.}
    \label{fig:detection_statistics}
\end{figure}

\begin{figure}[!t]
    \centering
    \begin{tikzpicture}
        \begin{axis}[
            xlabel={Number of vehicles},
            ylabel={$\text{GOSPA}_M$},
            ymin=0, ymax=2.1,
            grid=both,
            width=7.5cm,
            height=5.5cm
        ]
        \addplot+[mark=*, error bars/.cd, y dir=both, y explicit]
        coordinates {
            (5,1.5400) +- (0,0.4610)
            (10,1.2694) +- (0,0.2525)
            (15,1.1335) +- (0,0.2123)
            (20,1.1013) +- (0,0.2207)
            (25,1.0145) +- (0,0.2173)
            (30,0.9324) +- (0,0.2132)
            (35,0.9050) +- (0,0.1154)
            (40,0.8158) +- (0,0.2043)
            (45,0.8407) +- (0,0.1485)
            (50,0.7802) +- (0,0.1274)
        };
        \end{axis}
    \end{tikzpicture}
    \caption{Evolution of the $\text{GOSPA}_M$ as fleet size increases. Error bars indicate $\pm 1$
    standard deviation.}
    \label{fig:gospam_vs_vehicles}
\end{figure}
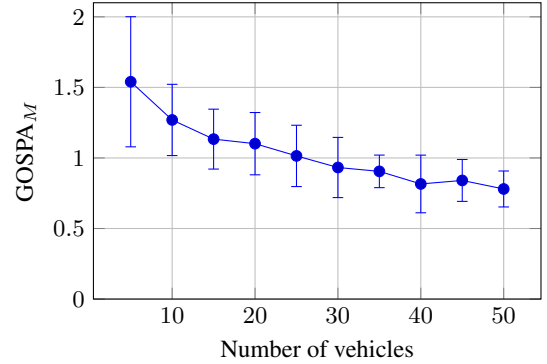

Fig.~\ref{fig:f1_vs_vehicles},   \ref{fig:detection_statistics} and \ref{fig:gospam_vs_vehicles} illustrate the impact of the fleet size on mapping performance. Together, these figures provide complementary perspectives on the influence of fleet size: the F1-score captures the detection performance,  the TP/FP/FN statistics provide detailed insight into the evolution of detection errors and the $\text{GOSPA}_M$ metric reflects the overall mapping quality.

The evolution of the F1-score is shown in Fig.~\ref{fig:f1_vs_vehicles}. The F1-score increases from approximately 0.75 with five vehicles to about 0.95 with more than 25 vehicles. This improvement reflects the increasing reliability of the aggregated detections as more observations become available. However, the curve appears to have plateaued toward the end. This might suggest that there are no further improvements in mapping performance. 

The detailed detection statistics are presented in Fig.~\ref{fig:detection_statistics}, which shows the evolution of true positives (TP), false positives (FP), and false negatives (FN). As the fleet size increases, the number of true positives slightly increases while both false positives and false negatives decrease. However, beyond roughly 30 vehicles these quantities stabilize, indicating that most traffic signs have already been sufficiently observed and confirmed.

Fig.~\ref{fig:gospam_vs_vehicles} shows the evolution of the $\text{GOSPA}_M$ as a function of the number of contributing vehicles. The results show a clear improvement in map quality when the first vehicles are added to the system. In particular, the $\text{GOSPA}_M$ value decreases significantly between 5 and 15 vehicles, reflecting improved localization accuracy and reduced detection errors. Beyond approximately 30 vehicles, the improvement becomes more gradual and the metric continues to decrease. This trend indicates that the spatial accuracy of the mapping continues to improve. This information is not visible in the other graphs. 

Another advantage of $\text{GOSPA}_M$ is that it is a measure for which a confidence interval can be calculated. It can be observed that once data from 10 vehicles has been collected, the uncertainty no longer changes significantly. 

Overall, these results highlight the diminishing returns characteristic of crowdsourced mapping systems. While increasing the number of contributing vehicles significantly improves map quality for small fleet sizes, additional vehicles provide progressively smaller gains once a sufficient observation density has been reached.

\subsection{Discussion}\label{Discussion}
The experimental study demonstrates that $\text{GOSPA}_M$ is a highly relevant measure for implementing crowdsourcing systems that utilize data collected by vehicles.  
In particular, it makes it possible to compare different approaches to processing the collected data and to select the one that results in the best mapping architecture. 
The experiments also confirm the importance of combining semantic grouping with spatial clustering when aggregating detections from multiple vehicles. Grouping detections by semantic type before spatial clustering prevents nearby signs of different types from being incorrectly merged, which is a frequent configuration in urban environments where multiple traffic signs may share the same pole or structure.

$\text{GOSPA}_M$ provides a convenient framework for studying the impact of various factors, such as fleet size, in order to collaboratively update a map of traffic signs.
Based on our results, the most significant improvements occur when the number of contributing vehicles increases from 5 to approximately 30 vehicles. Beyond this range, the improvement in mapping is limited to the accuracy of the location of the signs.
These findings provide useful insights for the deployment of crowdsourced mapping systems. While increasing the number of contributing vehicles improves map quality, the results show clear diminishing returns once a sufficient observation density has been reached. These results suggest that an optimal fleet size may exist depending on the needs of the application, where additional vehicles provide limited benefit relative to the associated data collection and processing costs.

Such insights are particularly relevant for connected vehicle fleets contributing to the validation and maintenance of digital maps used for ADAS or AD, where efficient data collection strategies are essential.

\section{Conclusion}\label{Conclusion}

This paper presented a fleet simulation system for crowdsourced traffic sign mapping that combines realistic sensor modeling with aggregation and comprehensive clustering algorithm evaluation. 
A systematic comparison of four clustering algorithms establishes DBSCAN with semantic-first filtering as the best approach. Numerous simulation tests were carried out on a route traveled by an experimental vehicle using a real HD map within a limited area corresponding to a real-world scenario. The results highlighted the utility of the $\text{GOSPA}_M$ in assessing the quality of crowdsourced mapping. In the considered experimental scenario, the results suggest that approximately 30 journeys by different vehicles appear to be sufficient to achieve high-quality traffic sign mapping within the considered map tile.

This work opens up many avenues for further research.
Evaluating the proposed approach on more diverse road environments (urban, highway, rural) would help assess the generality of the observed trends.
Future work will also investigate mixed fleets combining vehicles equipped with high-end perception systems and others with more limited sensing capabilities, in order to better understand the contribution of heterogeneous sensing sources to crowdsourced mapping. 

\section{Acknowledgments}

The authors acknowledge OpenStreetMap contributors for providing map data used in this study. These data were used as map sources to be evaluated and were not considered as ground truth. Generative AI tools were used for coding assistance and language improvement. All AI-assisted content was reviewed and validated by the authors, who remain fully responsible for the content of this work.

\bibliography{references_isprs}

\end{document}